\documentclass[11pt]{article}

\usepackage[utf8]{inputenc}
\usepackage[margin=1in]{geometry}
\usepackage{amsmath,amssymb,amsthm}
\usepackage{graphicx}
\usepackage{hyperref}
\usepackage{natbib}
\usepackage{booktabs}
\usepackage{algorithm}
\usepackage{algorithmic}
\usepackage{xcolor}

\hypersetup{
    colorlinks=true,
    linkcolor=blue,
    filecolor=magenta,      
    urlcolor=cyan,
    citecolor=blue,
}

\title{Low-Dimensional Execution Manifolds in Transformer Learning Dynamics: Evidence from Modular Arithmetic Tasks}

\author{
    Yongzhong Xu%
    \thanks{abbyxu@gmail.com. Code available at \url{https://github.com/skydancerosel/bubble-modadd}}
}

\date{}

\begin{document}

\maketitle

\begin{abstract}
We investigate the geometric structure of learning dynamics in overparameterized transformer models through carefully controlled modular arithmetic tasks. Our primary finding is that despite operating in high-dimensional parameter spaces ($d=128$), transformer training trajectories rapidly collapse onto low-dimensional execution manifolds of dimension $3$--$4$. This dimensional collapse is robust across random seeds and moderate task difficulties, though the orientation of the manifold in parameter space varies between runs. We demonstrate that this geometric structure underlies several empirically observed phenomena: (1) sharp attention concentration emerges as saturation along routing coordinates within the execution manifold, (2) SGD commutators are preferentially aligned with the execution subspace (up to $10\times$ random baseline) early in training, with $>92\%$ of non-commutativity confined to orthogonal staging directions and this alignment decreasing as training converges, and (3) sparse autoencoders capture auxiliary routing structure but fail to isolate execution itself, which remains distributed across the low-dimensional manifold. Our results suggest a unifying geometric framework for understanding transformer learning, where the vast majority of parameters serve to absorb optimization interference while core computation occurs in a dramatically reduced subspace. These findings have implications for interpretability, training curriculum design, and understanding the role of overparameterization in neural network learning.
\end{abstract}

\section{Introduction}

Transformer models have demonstrated remarkable capabilities across diverse domains \citep{vaswani2017attention}, yet fundamental questions about their learning dynamics remain open. Empirical observations---including sharp attention concentration (``bubbling'') \citep{elhage2021mathematical}, grokking-like generalization transitions \citep{power2022grokking}, interpretable circuit formation \citep{olah2020zoom}, and surprising robustness to noisy optimization---are typically studied in isolation. This work proposes a unifying geometric explanation: training dynamics in overparameterized transformers rapidly collapse onto low-dimensional \emph{execution manifolds}, and many observed phenomena emerge as natural projections or consequences of this severe dimensional reduction.

Traditional analyses of neural network learning focus on the final trained model as a static function. In contrast, we study the \emph{trajectory of learning itself}---the geometry and dynamics of the path through parameter space. Our approach is motivated by the observation that in highly overparameterized networks, not all dimensions of parameter space contribute equally to task performance. We hypothesize that learning concentrates along a small number of critical directions, with the remaining dimensions serving auxiliary roles in optimization.

To test this hypothesis rigorously, we employ a carefully controlled experimental paradigm: marker-based modular arithmetic. This task provides sufficient complexity to engage transformer mechanisms (attention, value routing, composition) while remaining tractable for detailed geometric analysis. By varying task difficulty through the number of markers and curriculum structure, we can probe how learning geometry responds to computational demands.

\subsection{Contributions}

Our main contributions are:

\begin{itemize}
    \item \textbf{Discovery of execution manifolds:} We demonstrate that attention-only transformer training trajectories collapse onto $3$--$4$ dimensional subspaces despite $d=128$ parameter dimensions, with dimension stable across random seeds.
    
    \item \textbf{Geometric explanation of attention bubbling:} Sharp attention concentration emerges naturally as saturation along routing coordinates within the execution manifold, providing a continuous geometric interpretation of a previously discrete-seeming phenomenon.
    
    \item \textbf{Localization of SGD non-integrability:} SGD commutators $[\nabla_A, \nabla_B] = \theta_{AB} - \theta_{BA}$ are large in ambient space. When projected onto the execution subspace (constructed from PCA of the training trajectory), the execution basis captures $2$--$10\times$ more commutator energy than a random subspace of equal dimension, with this ratio decreasing over training as non-commutativity progressively rotates out of the execution manifold. The perpendicular component accounts for $>92\%$ of total commutator magnitude throughout.
    
    \item \textbf{Execution-routing separation:} Sparse autoencoders capture auxiliary routing structure but do not isolate execution, which remains distributed across the low-dimensional manifold.
    
    \item \textbf{Curriculum effects on geometry:} Mixed-task training discovers unified execution manifolds enabling compositional generalization, while curriculum learning leads to catastrophic forgetting via path-dependent movement through fragmented solution regions.
\end{itemize}

\section{Related Work}

\textbf{Mechanistic interpretability.} Recent work has made progress in understanding transformer circuits through techniques like activation patching \citep{elhage2021mathematical}, attention pattern analysis, and sparse dictionary learning \citep{cunningham2023sparse}. Our work complements these approaches by focusing on the geometric structure of parameter-space trajectories rather than activation-space features.

\textbf{Grokking and generalization.} The phenomenon of delayed generalization (``grokking'') has been observed in modular arithmetic tasks \citep{power2022grokking}. We provide a geometric interpretation: grokking may correspond to discovering the low-dimensional integrable subspace where compositional operations commute.

\textbf{Loss landscape geometry.} Extensive work has studied loss landscape structure \citep{li2018visualizing}, mode connectivity \citep{garipov2018loss}, and solution manifolds \citep{fort2019emergent}. We extend this by analyzing \emph{training trajectory} geometry rather than just final solution geometry.

\textbf{Overparameterization theory.} Theoretical work on overparameterized networks \citep{allen2019convergence,jacot2018neural} has focused primarily on convergence guarantees. We provide an empirical perspective on the \emph{functional role} of excess dimensions: absorbing optimization interference while computation proceeds in a low-dimensional subspace.

\section{Methods}

\subsection{Task Design: Marker-Based Modular Arithmetic}

We design a synthetic task that isolates key transformer capabilities while enabling precise geometric analysis. Each input sequence has length $T = 32$ and contains $m$ non-adjacent marked positions indicated by special marker tokens. Each marker is immediately followed by a value token drawn from $\{0, 1, \ldots, C-1\}$. The model's task is to compute the sum of all marked values modulo $C$. Remaining positions contain i.i.d. distractor tokens that must be ignored.

\textbf{Task requirements.} This task requires three core competencies:
\begin{itemize}
    \item \textbf{Selective attention:} identifying and attending to marker tokens while ignoring distractors
    \item \textbf{Value routing:} extracting and aggregating the values following marked positions
    \item \textbf{Compositional arithmetic:} computing the modular sum across an arbitrary number of terms
\end{itemize}

\textbf{Difficulty control.} Task difficulty is controlled via:
\begin{itemize}
    \item Number of markers $m$ (ranging from $1$ to $6$ in our experiments)
    \item Modulus $C$ (set to $8$ or $16$ depending on experiment)
    \item Training curriculum (mixed simultaneous training versus progressive increase in $m$)
\end{itemize}

This parametric control allows us to systematically probe how computational complexity affects learning geometry.

\subsection{Model Architecture}

We study two architectural variants to isolate the role of different components:

\textbf{Attention-only transformers.} Multi-layer models with self-attention mechanisms but no MLP (multi-layer perceptron) blocks. These models rely purely on attention for routing and computation. For an attention-only model with $L$ layers and embedding dimension $d$, each layer contains four weight matrices: $W_Q, W_K, W_V, W_O \in \mathbb{R}^{d \times d}$.

\textbf{Standard transformers.} Models with both attention and MLP components in each block, following conventional transformer architecture.

All models use embedding dimension $d = 128$ across all layers. For attention-only models, we focus on the four attention weight matrices which together constitute the complete parameterization. This architectural simplicity enables clean geometric analysis of parameter trajectories without confounding effects from MLP components.

\textbf{Training.} We use standard stochastic gradient descent with learning rate $\eta = 0.001$ and batch size $64$. We track complete parameter trajectories by saving checkpoints at regular intervals (every $100$ steps) throughout training. Multiple random initializations ( across $20$ seeds) are trained for each experimental condition to assess consistency of geometric findings.

\subsection{Geometric Analysis Framework}

\subsubsection{Effective Rank and Intrinsic Dimension}
\label{sec:effective_rank}

To quantify the dimensional structure of learning trajectories, we employ principal component analysis (PCA) on parameter sequences. For a given attention weight matrix $W$ evolving over time, we collect snapshots $W(t_1), W(t_2), \ldots, W(t_n)$ and flatten them into vectors $\vec{w}_i \in \mathbb{R}^{d^2}$.

We compute the covariance matrix:
\begin{equation}
\Sigma = \frac{1}{n} \sum_{i=1}^n (\vec{w}_i - \bar{\vec{w}})(\vec{w}_i - \bar{\vec{w}})^T
\end{equation}
and its eigendecomposition $\Sigma = V \Lambda V^T$ where $\Lambda = \text{diag}(\lambda_1, \ldots, \lambda_{d^2})$ with $\lambda_1 \geq \lambda_2 \geq \cdots \geq \lambda_{d^2} \geq 0$.

The \textbf{effective rank} is defined as the number of principal components needed to capture $90\%$ of the variance:
\begin{equation}
r_{90} = \min\left\{k : \frac{\sum_{i=1}^k \lambda_i}{\sum_{i=1}^{d^2} \lambda_i} \geq 0.90\right\}
\end{equation}

This analysis is basis-invariant: the intrinsic dimension of the execution manifold is a geometric property independent of parameter coordinates.

\subsubsection{SGD Commutator Analysis}

To analyze whether SGD exhibits integrable dynamics, we compute \textbf{commutators}:
\begin{equation}
[\nabla_A, \nabla_B] := \theta_{AB} - \theta_{BA}
\end{equation}
where $\theta_{AB}$ represents the parameter value after applying gradients from independent minibatches $A$ then $B$:
\begin{align}
\theta_A &= \theta_0 - \eta \nabla_{\mathcal{L}}(\theta_0; A) \\
\theta_{AB} &= \theta_A - \eta \nabla_{\mathcal{L}}(\theta_A; B)
\end{align}
and similarly for $\theta_{BA}$.

For integrable dynamics, commutators vanish; significant commutator norms indicate path-dependent, non-integrable optimization.

\textbf{Projection decomposition.} Let $B \in \mathbb{R}^{P \times K}$ be an orthonormal basis for the learned execution subspace, constructed from the top $K$ principal components of each attention block's weight \emph{trajectory} (Section~\ref{sec:effective_rank}), embedded in the full parameter space and QR-orthonormalized. The projection $\Pi = BB^T$ decomposes commutators:
\begin{align}
[\nabla_A, \nabla_B] &= \Pi([\nabla_A, \nabla_B]) + (I - \Pi)([\nabla_A, \nabla_B]) \\
&= [\nabla_A, \nabla_B]_{\parallel} + [\nabla_A, \nabla_B]_{\perp}
\end{align}

The projection fraction $\rho_{\mathrm{exec}} = \|[\nabla_A, \nabla_B]_{\parallel}\| / \|[\nabla_A, \nabla_B]\|$ quantifies what fraction of non-commutativity lives within the execution manifold.

\textbf{Scale normalization.} We report both the raw commutator $\delta = \theta_{AB} - \theta_{BA}$ and the \emph{normalized defect}:
\begin{equation}
D = \frac{\|\delta\|}{\|\eta\,\nabla_{\mathcal{L}}(\theta_0; A)\| \cdot \|\eta\,\nabla_{\mathcal{L}}(\theta_0; B)\|}
\end{equation}
which measures non-commutativity relative to step magnitude.

\textbf{Random baseline control.} To distinguish genuine geometric alignment from a trivial dimensionality artifact ($K$ dimensions out of $P$ will capture $\sim\!\sqrt{K/P}$ of any random vector), we compare $\rho_{\mathrm{exec}}$ against $\rho_{\mathrm{rand}}$: the projection fraction obtained from a random $K$-dimensional orthonormal basis (averaged over 10 trials). The ratio $\rho_{\mathrm{exec}} / \rho_{\mathrm{rand}}$ is our key diagnostic; values significantly above $1.0$ indicate that the execution subspace captures more commutator energy than expected by chance.

\subsection{Sparse Autoencoder Probing}

To investigate whether learned representations can be decomposed into interpretable features, we train sparse autoencoders (SAEs) on intermediate activations. The SAE architecture consists of:
\begin{itemize}
    \item Encoder: $h = \text{ReLU}(W_e x + b_e)$
    \item Decoder: $\hat{x} = W_d h + b_d$
\end{itemize}

Training minimizes:
\begin{equation}
\mathcal{L}_{\text{SAE}} = \|\hat{x} - x\|^2 + \lambda \|h\|_1
\end{equation}
where $\lambda$ controls sparsity.

After training, we perform \textbf{targeted ablation}: systematically zeroing individual SAE latents and measuring the effect on task accuracy. This probes whether specific latents correspond to interpretable computational components.

\section{Results}

\subsection{Dimensional Collapse onto Execution Manifolds}

Our central finding is that attention parameters in attention-only transformers undergo rapid dimensional collapse during training. Despite the high ambient dimensionality ($d = 128$ for each attention matrix, giving $d^2 = 16{,}384$ parameters per matrix), parameter trajectories concentrate onto subspaces of dimension $3$--$4$ within the first $20$--$30\%$ of training steps.

\begin{figure}[t]
\centering
\includegraphics[width=0.7\textwidth]{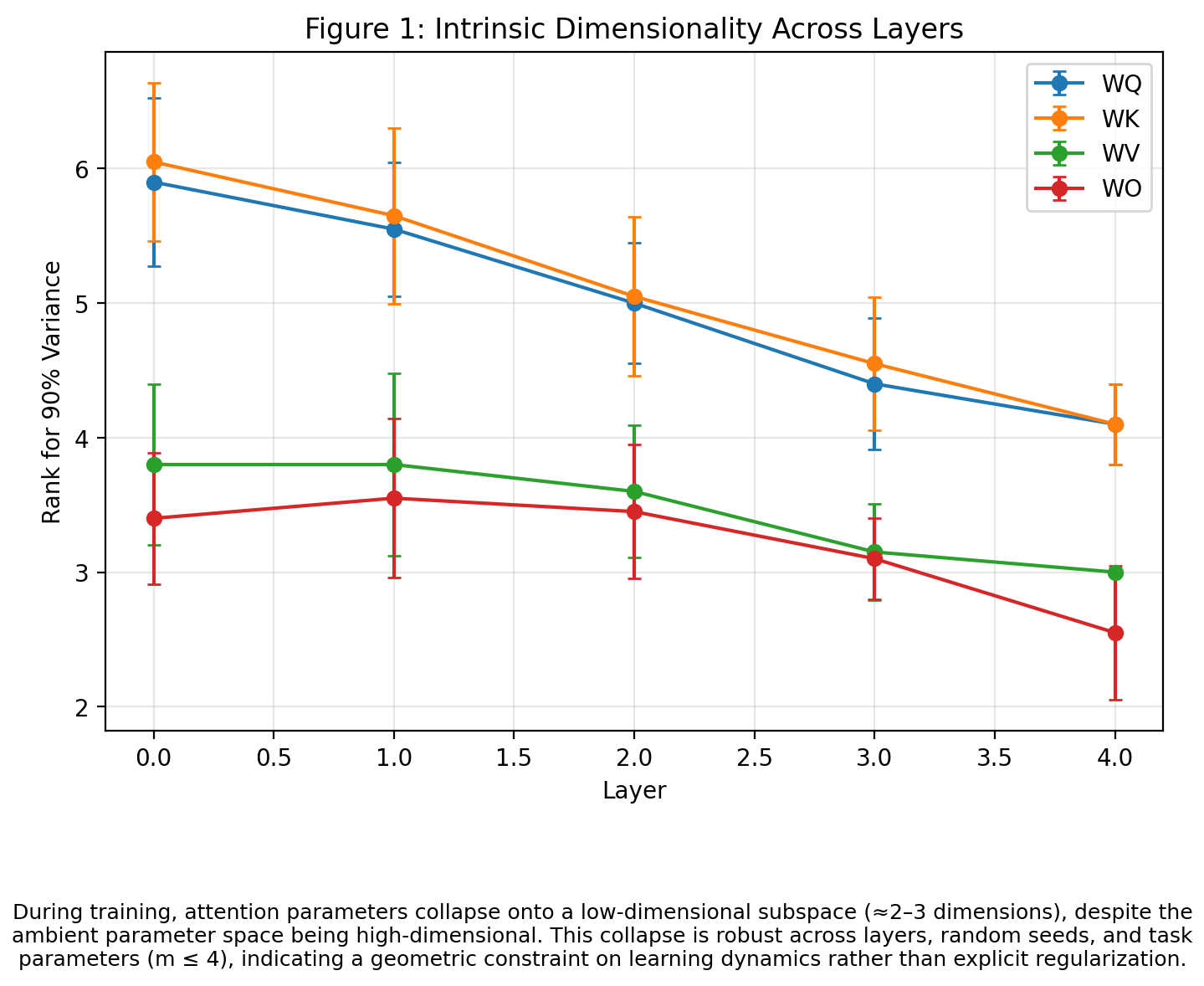}
\caption{Effective rank (at $90\%$ variance) of attention weight matrices ($W_Q$, $W_K$, $W_V$, $W_O$) across layers. All matrices collapse to dimension $3$--$5$ despite ambient dimension $d^2 = 16{,}384$, with $W_V$ and $W_O$ consistently lower-dimensional than $W_Q$ and $W_K$. Error bars show standard deviation across 5 random seeds.}
\label{fig:intrinsic_dim}
\end{figure}

\textbf{Consistency across components.} Figure~\ref{fig:intrinsic_dim} shows the effective rank of $W_Q$, $W_K$, $W_V$, and $W_O$ matrices across training for a representative run with $m = 4$ markers. All four attention matrices converge to effective ranks in the range $3$--$5$, with $W_V$ and $W_O$ consistently lower-dimensional than $W_Q$ and $W_K$. This demonstrates that dimensional collapse is not specific to a single component but represents a network-wide phenomenon. The consistency across layers further indicates that dimensional reduction is a fundamental property of how these models learn the task.

\textbf{Seed consistency and orientation variance.} While the intrinsic dimension ($3$--$4$) is consistent across random seeds, the orientation of this subspace in raw parameter coordinates is seed-dependent. Different initializations discover different rotations of functionally equivalent solutions. This suggests that the execution manifold is an intrinsic geometric object, not an artifact of particular coordinate systems or initialization schemes.

\textbf{Task complexity scaling.} The dimensionality of the execution manifold remains stable for moderate task complexity ($m \leq 6$). However, preliminary experiments suggest that extremely difficult variants or fundamentally different operations may require slightly higher-dimensional manifolds ($4$--$6$ dimensions for modular multiplication versus $3$--$4$ for addition), indicating that manifold dimension may scale predictably with computational complexity.

\subsection{Attention Bubbling as Geometric Saturation}

Sharp attention concentration---the phenomenon of attention weights forming narrow ``bubbles'' focused on specific tokens---emerges naturally as a consequence of the low-dimensional geometry. Rather than being a discrete architectural feature, bubbling represents continuous saturation along routing coordinates within the execution manifold.

\begin{figure}[t]
\centering
\includegraphics[width=0.7\textwidth]{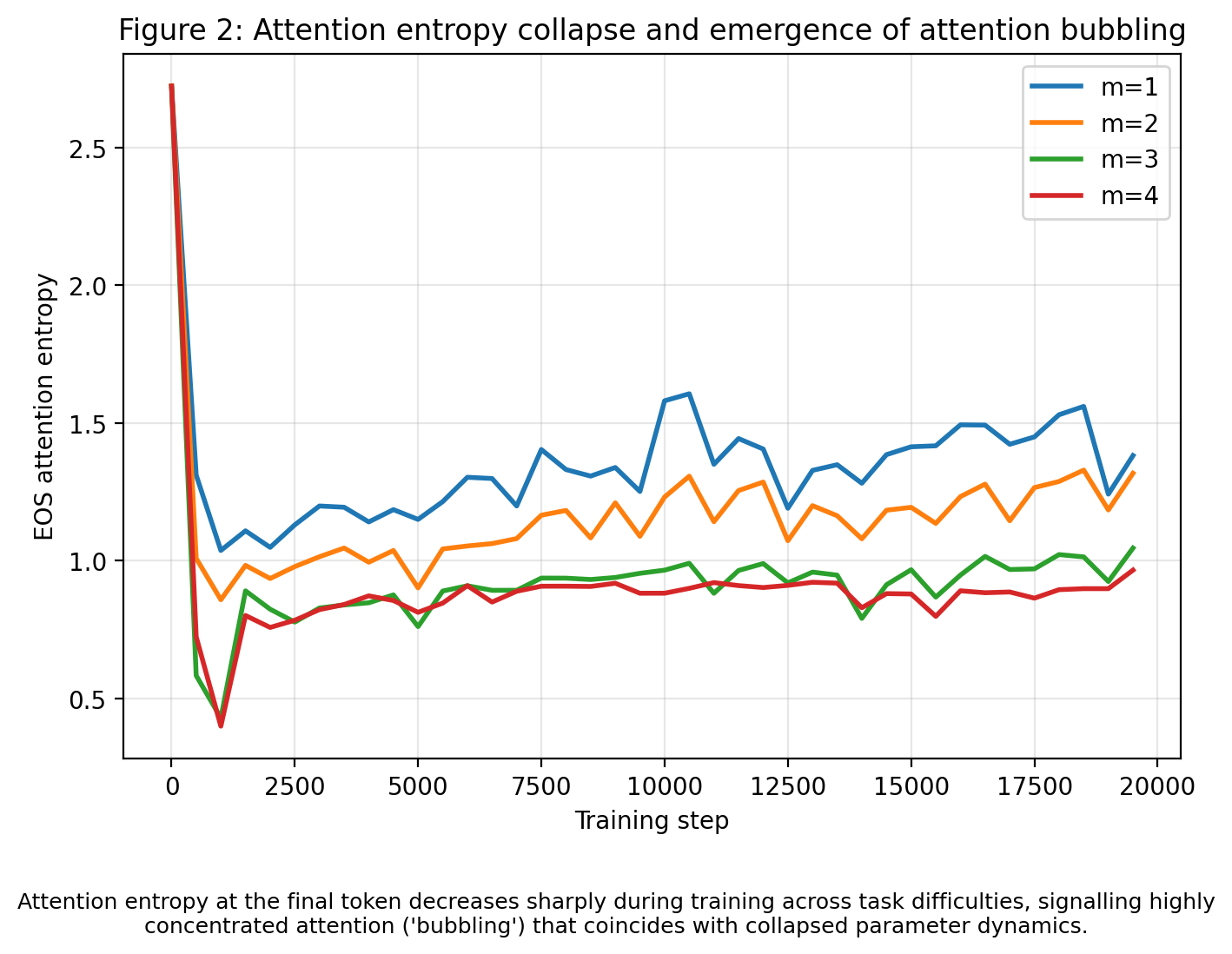}
\caption{Mean attention entropy decreases as training progresses, indicating the emergence of sharp attention ``bubbles.'' The decrease is continuous and smooth, consistent with geometric saturation along the execution manifold.}
\label{fig:entropy}
\end{figure}

\textbf{Entropy dynamics.} Figure~\ref{fig:entropy} displays attention entropy over training:
\begin{equation}
H(A) = -\sum_{j} A_{ij} \log A_{ij}
\end{equation}
where $A_{ij}$ is the attention weight from token $i$ to token $j$.

Initially, attention distributions are relatively uniform (high entropy). As training progresses and parameters move along the execution manifold, entropy drops sharply, indicating concentration onto specific tokens. This transition is smooth rather than abrupt, consistent with continuous geometric movement rather than a discrete phase transition.

\textbf{Correlation with manifold structure.} The timing of bubble formation correlates with the stabilization of the execution manifold structure. Once parameters have collapsed onto the $3$--$4$ dimensional subspace, further movement along this manifold drives attention toward increasingly sharp distributions. This interpretation unifies attention bubbling with the overall geometric picture: both are manifestations of learning dynamics constrained to a low-dimensional solution space.

\subsection{Localization of SGD Non-Integrability}

\begin{figure}[t]
\centering
\includegraphics[width=0.7\textwidth]{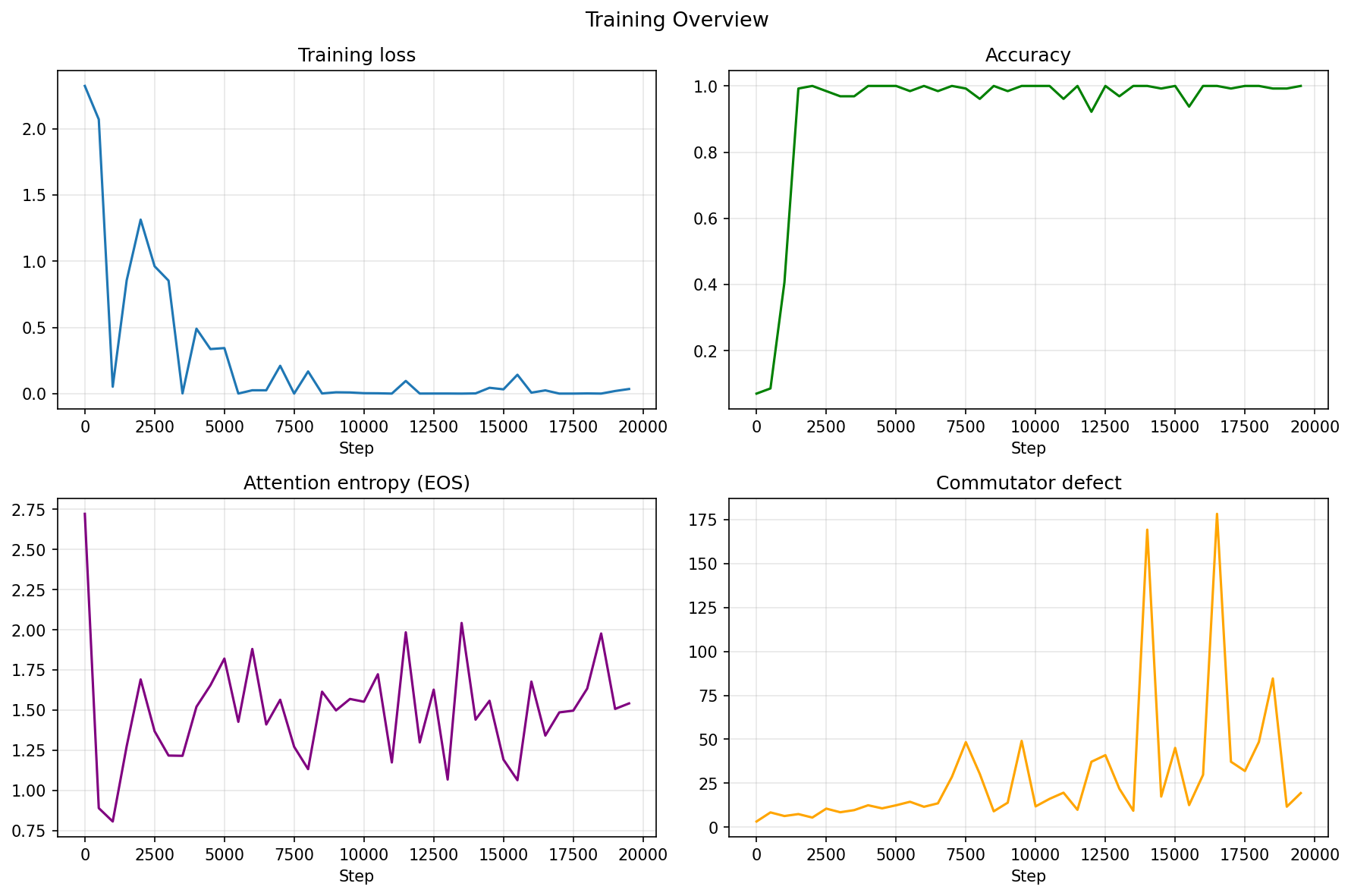}
\caption{Training overview. Top-left: training loss converges by step $\sim\!5000$. Top-right: accuracy reaches $100\%$ and remains stable. Bottom-left: attention entropy (at EOS token) drops sharply early in training, reflecting bubble formation. Bottom-right: normalized commutator defect $D = \|\delta\| / (\|\eta g_A\| \cdot \|\eta g_B\|)$ \emph{increases} throughout training despite loss convergence, with intermittent spikes reaching $100$--$175\times$ step magnitude, indicating persistent non-commutativity in the full parameter space.}
\label{fig:commutator_raw}
\end{figure}

\begin{figure}[t]
\centering
\includegraphics[width=0.7\textwidth]{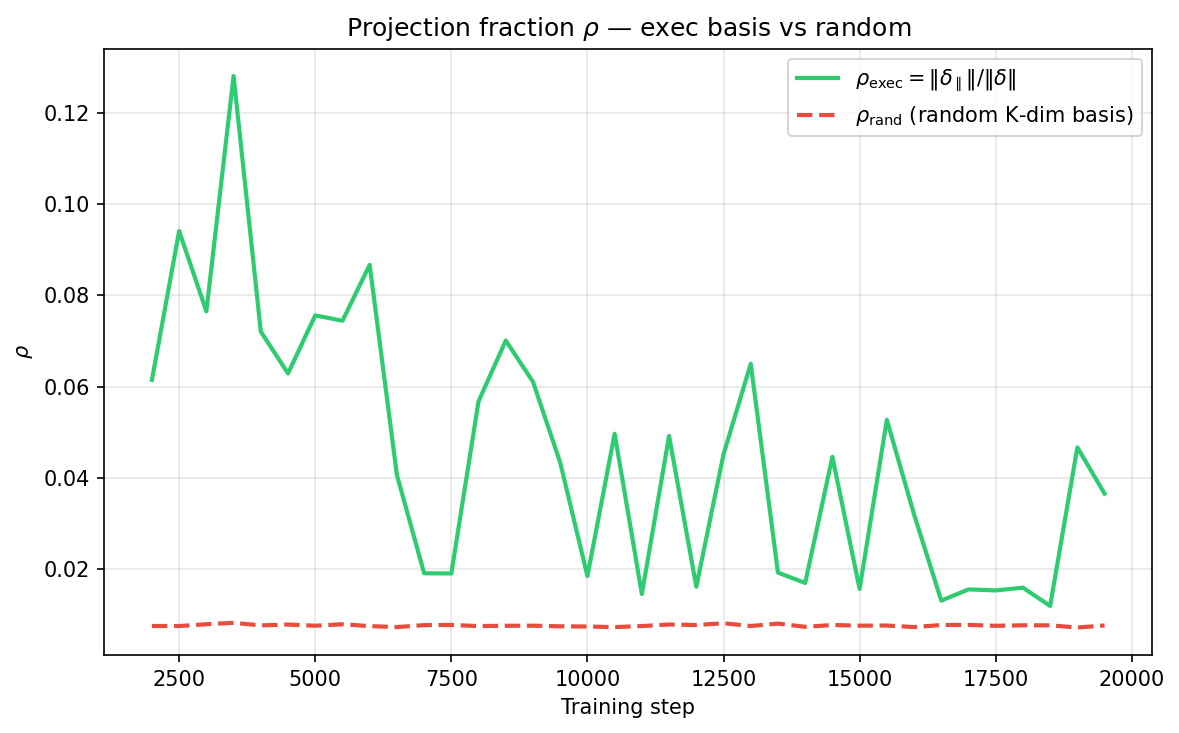}
\caption{Projection fraction $\rho_{\mathrm{exec}} = \|\delta_\parallel\|/\|\delta\|$ (green, solid) versus random baseline $\rho_{\mathrm{rand}}$ (red, dashed) over training. The execution subspace is constructed from PCA of the weight trajectory (Section~\ref{sec:effective_rank}); the random baseline averages over 10 random $K$-dimensional orthonormal bases of matching dimension. The execution basis consistently captures $2$--$10\times$ more commutator energy than random, confirming structured geometric alignment rather than a trivial dimensionality artifact.}
\label{fig:comm_ratio}
\end{figure}

\begin{figure}[t]
\centering
\includegraphics[width=\textwidth]{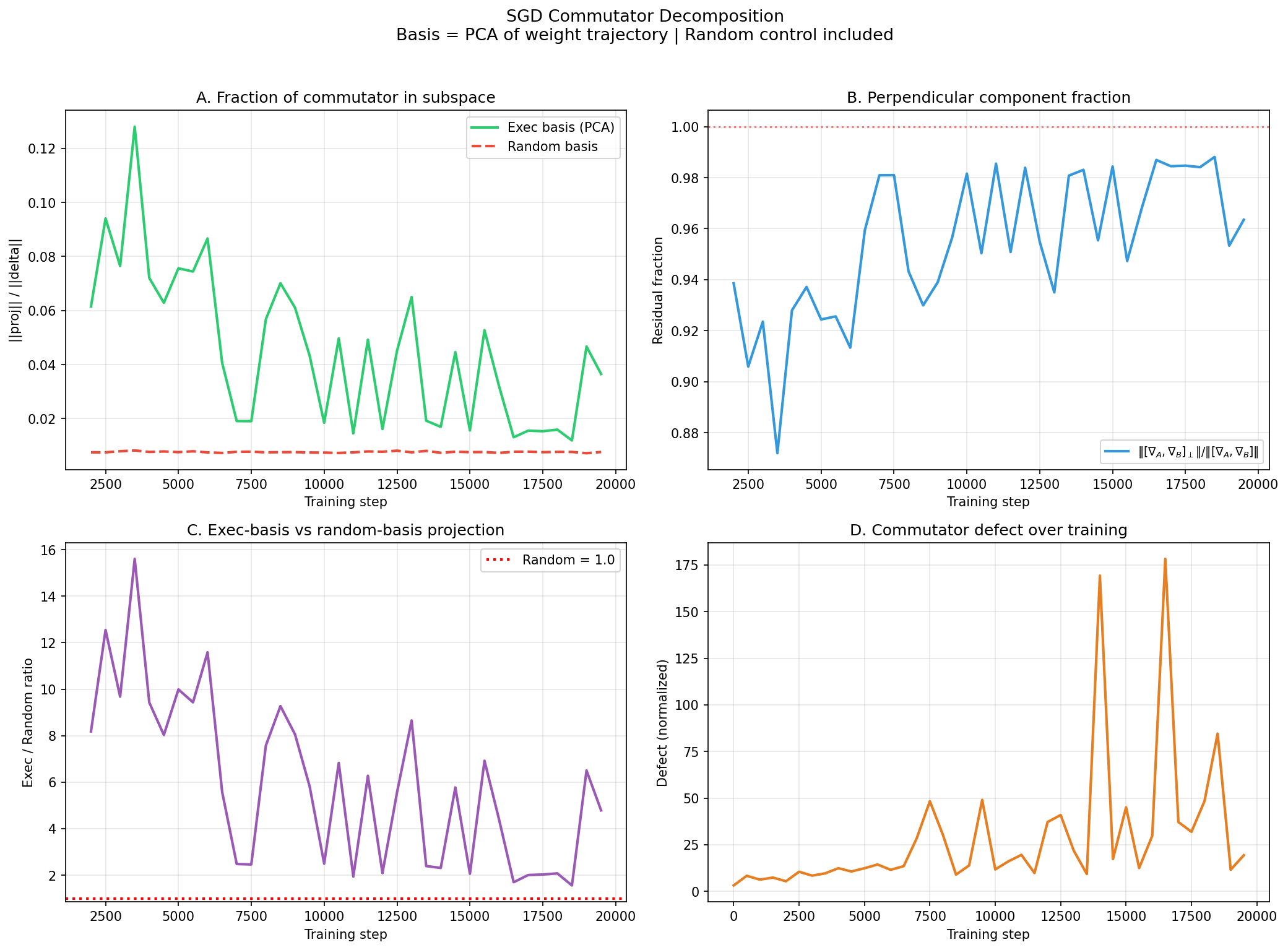}
\caption{Corrected commutator decomposition (PCA trajectory basis with random control). \textbf{A}: Projection fractions $\rho_{\mathrm{exec}}$ (green) and $\rho_{\mathrm{rand}}$ (red, dashed); the execution basis captures a small but significantly above-random fraction of commutator energy. \textbf{B}: Perpendicular component fraction $\|\delta_\perp\|/\|\delta\|$ (blue); $>92\%$ throughout training, rising to $>98\%$ late. \textbf{C}: Exec/random ratio $\rho_{\mathrm{exec}}/\rho_{\mathrm{rand}}$ (purple); peaks at $\sim\!14\times$ early, decays to $\sim\!2\times$ late, with the dashed line at $1.0$ marking the random expectation. \textbf{D}: Raw normalized defect $D$ (orange); grows throughout training, confirming that non-commutativity intensifies even after convergence.}
\label{fig:comm_decomp}
\end{figure}

SGD updates in the full high-dimensional parameter space exhibit strong non-commutativity. Figure~\ref{fig:commutator_raw} (bottom-right panel) shows the normalized commutator defect $D$ throughout training. Despite loss convergence by step $\sim\!5000$, the defect continues to grow with intermittent spikes, indicating that non-commutativity is not simply a manifestation of optimization difficulty but a persistent feature of the learning dynamics.

\textbf{Projection reveals structure.} To test whether this non-commutativity has geometric structure, we project commutators onto the learned execution subspace built from PCA of the weight trajectory (Section~\ref{sec:effective_rank}). Concretely, for each measurement step, we:
\begin{enumerate}
    \item Construct the execution basis $B \in \mathbb{R}^{P \times K}$ from the top-$K$ PCA components of the accumulated weight trajectory, QR-orthonormalized;
    \item Compute $\rho_{\mathrm{exec}} = \|B B^T \delta\| / \|\delta\|$, the fraction of the commutator $\delta = \theta_{AB} - \theta_{BA}$ captured by the execution subspace;
    \item Compute $\rho_{\mathrm{rand}}$ by averaging the same projection fraction over $10$ random $K$-dimensional orthonormal bases.
\end{enumerate}

Figure~\ref{fig:comm_ratio} shows the result: $\rho_{\mathrm{exec}}$ (green) is consistently small ($\approx 0.02$--$0.13$) but systematically above $\rho_{\mathrm{rand}}$ (red, dashed), confirming that the vast majority ($>92\%$) of non-commutativity lives in orthogonal staging directions, while the fraction within execution directions is small but geometrically structured.

\textbf{Comparison against random baseline.} The raw projection fraction alone does not distinguish genuine geometric alignment from a trivial dimensionality effect: \emph{any} $K$-dimensional subspace of a $P$-dimensional space captures $\sim\!\sqrt{K/P}$ of a random vector's norm. The critical diagnostic is the ratio $\rho_{\mathrm{exec}} / \rho_{\mathrm{rand}}$ (Figure~\ref{fig:comm_decomp}, panel~C). Early in training (first 20\%), this ratio reaches $\mathbf{9.7\times}$ (median): the execution subspace captures nearly $10\times$ more commutator energy than a random subspace of equal dimension. Late in training (last 20\%), the ratio decreases to $\mathbf{2.1\times}$, indicating that non-commutativity progressively \emph{leaves} the execution manifold as the model converges.

This temporal pattern---commutators initially aligned with execution directions, then rotating away---has a natural interpretation. During early learning, gradient updates are shaped by the task structure and therefore align with the execution manifold. As the model converges, the residual non-commutativity becomes increasingly orthogonal, reflecting optimization interference in staging dimensions that does not affect the learned computation.

\textbf{Geometric role of overparameterization.} The localization of non-integrability to orthogonal directions ($>92\%$ perpendicular throughout, rising to $>98\%$ late in training; Figure~\ref{fig:comm_decomp}, panel~B) has implications for the functional role of excess parameters. Extra dimensions provide a ``buffer space'' where optimization noise, task interference, and stochastic fluctuations can be sequestered without disrupting the core computational trajectory. The execution manifold acts as a stable attractor along which learning progresses, with the decreasing exec/random ratio indicating that residual non-commutativity is progressively expelled from execution directions as the model converges.

\subsection{Sparse Autoencoders and the Execution-Routing Distinction}

Training sparse autoencoders on intermediate activations reveals interpretable structure, but this structure is distinct from the execution manifold itself. A small number of SAE latents (typically $3$--$5$ out of several hundred) correlate strongly with task-relevant features such as the count of markers $m$ or positions of marked tokens.

\begin{figure}[t]
\centering
\includegraphics[width=0.7\textwidth]{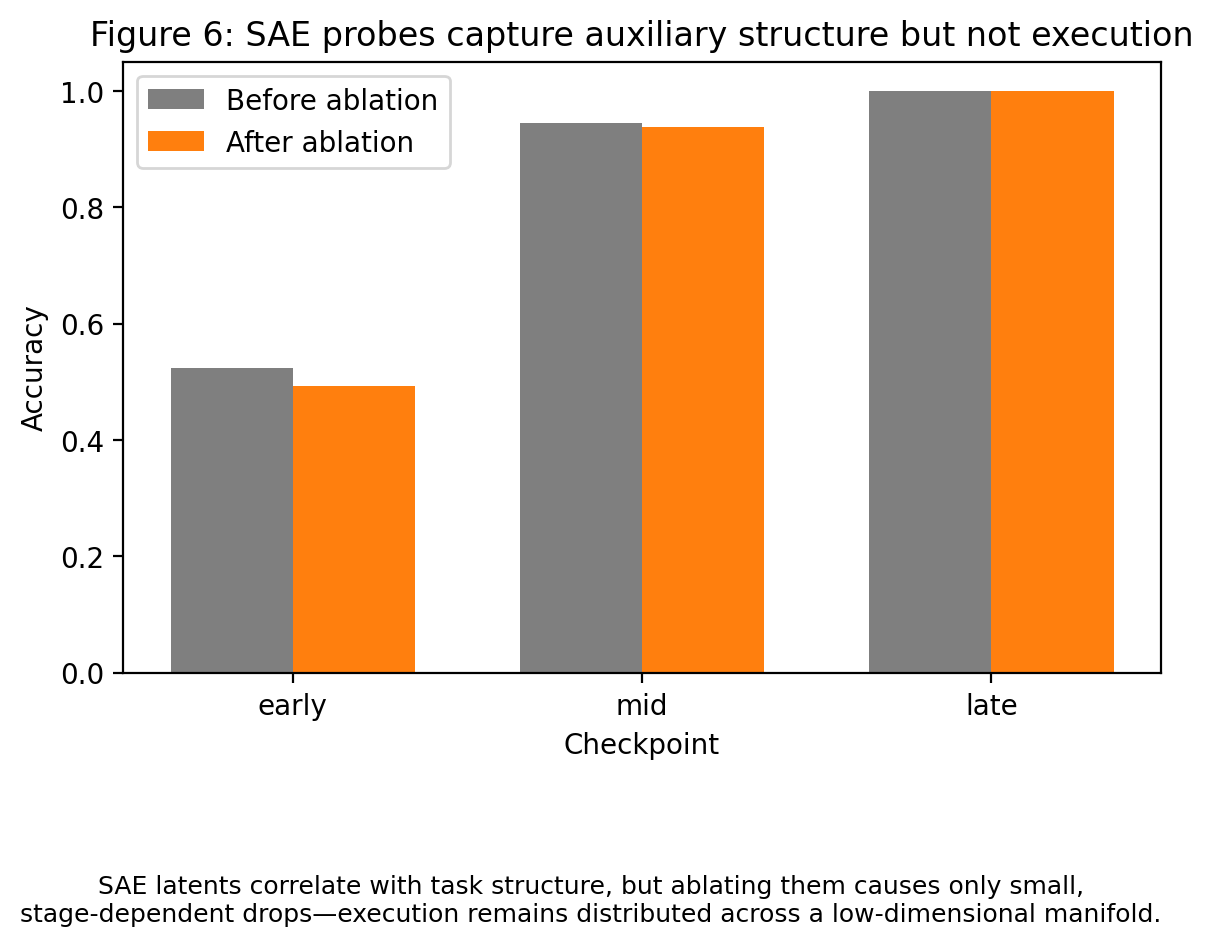}
\caption{Accuracy before and after ablating task-correlated SAE latents at three training stages. Ablation impact is uniformly small: the largest drop occurs early (when baseline accuracy is low), while mid- and late-training ablation is negligible. This indicates SAE features capture peripheral correlates rather than core execution, which remains distributed across the low-dimensional manifold.}
\label{fig:sae_ablation}
\end{figure}

\textbf{Uniformly small ablation impact.} Figure~\ref{fig:sae_ablation} shows ablation results: zeroing these task-correlated latents produces only minimal accuracy drops across all training stages. The largest effect occurs early in training (when the model has not yet learned robust representations and baseline accuracy is low), while mid- and late-training ablation has negligible impact. This pattern indicates that although SAE latents correlate with task structure, they do not causally contribute to core execution at any stage of learning.

\textbf{Execution remains distributed.} Crucially, SAE latents do not isolate execution itself. The core computation---modular addition over selected values---remains distributed across the low-dimensional execution manifold rather than localized to individual interpretable features. The near-zero ablation sensitivity even mid-training (when the model is already performing well) underscores this point: sparse features capture peripheral correlates of the task---such as marker counting or positional bookkeeping---but the fundamental execution geometry is not sparse-feature-decomposable in the SAE sense.

\subsection{Architectural and Curriculum Effects}

\textbf{MLP disruption of low-dimensional structure.} Adding MLP layers fundamentally disrupts the low-dimensional geometric picture. Standard transformer models with both attention and MLP components exhibit higher-dimensional training dynamics, with no clear collapse onto a $3$--$4$ dimensional subspace. Correspondingly, generalization performance degrades under comparable training budgets, suggesting that the clean geometric structure of attention-only models facilitates efficient learning.

\textbf{Curriculum vs. mixed training.} Training curriculum exerts strong effects on solution geometry. \emph{Mixed training}---presenting all task difficulties ($m = 1, 2, 3, 4$) simultaneously---forces discovery of a unified execution manifold that supports compositional generalization. In contrast, strict \emph{curriculum learning} (progressively increasing $m$ from $1$ to $4$) leads to catastrophic forgetting: as training advances to larger $m$ values, performance on smaller $m$ degrades significantly.

This forgetting is particularly striking because the model has sufficient capacity to represent all task variants simultaneously---indeed, mixed training achieves this. The curriculum-induced forgetting arises from path-dependent movement along a low-dimensional solution manifold: early training on simple cases (small $m$) discovers local solutions that occupy specific regions of staging space. Advancing to harder cases requires moving to different regions, disrupting earlier solutions. Mixed training avoids this trap by exploring the geometry more broadly and discovering the integrable subspace where all task variants coexist.

\subsection{Extension to Modular Multiplication}

Preliminary experiments with modular multiplication confirm that the execution manifold framework extends beyond addition. Multiplication tasks converge to slightly higher-dimensional manifolds ($4$--$6$ dimensions versus $3$--$4$ for addition), with dimensionality scaling predictably with computational complexity. All attention matrices again operate in low-dimensional subspaces, and the same geometric principles apply: commutator localization to orthogonal directions, preferential alignment of non-commutativity with execution subspaces relative to random baselines, and curriculum effects. This suggests that manifold dimensionality may be a robust, task-dependent quantity that reflects inherent computational structure.

\section{Discussion}

\subsection{Theoretical Interpretation: Overparameterization and Parallel Exploration}

Our findings suggest a novel perspective on neural network learning in overparameterized regimes. Rather than viewing excess parameters as mere redundancy, we propose that high dimensionality enables \emph{parallel exploration} of solution space: the model can simultaneously maintain multiple candidate solutions in orthogonal ``staging'' directions without interference. As training progresses, these explorations consolidate onto a low-dimensional execution manifold representing the discovered computational strategy.

This interpretation connects to dynamical systems theory, where execution manifolds can be viewed as \emph{attractors} in the learning dynamics. The geometry of these attractors---their dimension, stability, and basin structure---determines learning behavior. From this perspective, phenomena like grokking may correspond to transitions where the system discovers and locks onto a low-dimensional integrable subspace, enabling the compositional operations characteristic of systematic generalization.

The commutator analysis reveals a nuanced temporal picture: early in training, non-commutativity is preferentially aligned with the execution manifold ($\rho_{\mathrm{exec}}/\rho_{\mathrm{rand}} \approx 10\times$), reflecting task-structured gradient interference. As training converges, this ratio decreases to $\approx 2\times$, indicating that residual non-commutativity progressively leaves the execution subspace. Throughout training, the perpendicular component accounts for $>92\%$ of total commutator magnitude. This localization provides a mechanism for stabilization: staging directions absorb the non-commutative aspects of SGD, allowing continued exploration without disrupting the core computational trajectory.

\subsection{Implications for Interpretability and AI Safety}

The execution manifold framework has direct implications for interpretability research. Our results suggest a temporal and spatial decomposition of interpretability targets:

\begin{itemize}
    \item \textbf{When to interpret:} Sparse features (as recovered by SAEs) are most relevant early and mid-training, when auxiliary routing structure is being established. Geometric structure dominates late training.
    
    \item \textbf{Where to interpret:} Rather than analyzing the full parameter space, interpretability efforts should focus on the execution manifold---the low-dimensional subspace where computation actually occurs.
    
    \item \textbf{How to train for interpretability:} Mixed-task training may be crucial for discovering compositional structure. Curriculum learning, while seemingly pedagogically sound, can trap models in fragmented solution regions that resist systematic generalization.
\end{itemize}

For AI safety, the \emph{scaling hypothesis} is critical: if powerful models trained on complex tasks exhibit similar low-dimensional execution manifolds, this offers hope for interpretability at scale. Rather than facing an exponentially growing parameter space, we may be able to focus on identifying and analyzing the relevant low-dimensional computational subspace. However, our preliminary results on modular multiplication suggest that manifold dimension may grow with task complexity, and the scaling relationship requires further investigation.

\subsection{Limitations and Future Directions}

Our study is limited to controlled synthetic tasks with clear ground-truth structure. Whether similar geometric phenomena occur in naturalistic tasks (language modeling, vision, etc.) remains an open question. The marker-based modular arithmetic task may over-emphasize low-dimensional solutions due to its linear structure; more complex tasks could exhibit fundamentally different geometry.

The role of architecture warrants deeper investigation. Our finding that MLPs disrupt low-dimensional structure suggests a tension between architectural expressivity and geometric simplicity. Understanding this tradeoff could inform architecture design for interpretability.

Several promising directions for future work emerge from our findings:

\begin{itemize}
    \item \textbf{Scaling laws for execution dimension:} How does manifold dimension scale with model size, task complexity, and dataset diversity? Can we predict execution dimension from task structure?
    
    \item \textbf{Early-training diagnostics:} Can geometric measurements early in training (subspace distance, effective rank) reliably predict eventual generalization success?
    
    \item \textbf{Nonlinear extensions:} For tasks without global linear structure, how should intrinsic dimension be measured? Do local manifolds patch together into a global geometric picture?
    
    \item \textbf{Circuit composition dynamics:} How do multiple micro-tasks discovered during training interact and compose over time? What determines whether they merge into stable reusable circuits versus remaining entangled?
\end{itemize}

\section{Conclusion}

We have presented evidence that transformer learning dynamics collapse onto low-dimensional execution manifolds, with dimension $3$--$4$ for modular addition tasks despite ambient parameter space dimension $128$. This geometric structure unifies several empirical phenomena: attention bubbling emerges as saturation along routing coordinates, SGD commutators are preferentially aligned with the execution subspace early in training ($\rho_{\mathrm{exec}}/\rho_{\mathrm{rand}} \approx 10\times$) before rotating into orthogonal directions as training converges ($\approx 2\times$), and sparse features capture auxiliary routing structure distinct from distributed execution.

The localization of most ($>92\%$) SGD non-commutativity to orthogonal staging directions, confirmed by comparison against random-subspace baselines, suggests a fundamental role for overparameterization: extra dimensions absorb optimization interference, allowing core computation to proceed via approximately path-independent updates. Training curriculum critically determines whether models discover unified execution manifolds supporting compositional generalization or become trapped in fragmented solution regions.

Our framework shifts the focus of neural network analysis from final trained models to the geometry of learning trajectories. Rather than asking ``what function does this network compute,'' we ask ``what geometric path did learning follow through parameter space.'' This perspective opens new avenues for understanding generalization, interpretability, and the role of architecture in shaping learning dynamics.

If similar geometric principles govern learning in larger, more complex models, the execution manifold framework may provide a path toward interpretability at scale: by identifying and analyzing the relevant low-dimensional computational subspace rather than confronting the full exponentially-large parameter space. Testing this scaling hypothesis is a critical direction for future research with implications for both understanding and aligning advanced AI systems.

\section*{Acknowledgments}

This work was primarily inspired by two lines of research: the mechanistic interpretability program at Anthropic, particularly the transformer circuits framework \citep{elhage2021mathematical,olah2020zoom} and the sparse autoencoder investigations into monosemanticity \citep{bricken2023monosemanticity}; and the scaling analysis of sparse autoencoders at OpenAI \citep{gao2024scaling}. The geometric perspective---especially the analogy between dimensional collapse in learning dynamics and energy concentration in variational problems---draws on the author's earlier work in nonlinear PDE and contact geometry, particularly the bubble analysis and concentration-compactness tradition, the theory of critical points at infinity in the sense of Bahri, and the infinite-dimensional Morse-theoretic ideas underlying Floer homology and Hofer's work on the Weinstein conjecture. Code and experimental data are available at \url{https://github.com/skydancerosel/bubble-modadd}.

\bibliographystyle{plainnat}
\bibliography{references}

\end{document}